%% file: main.tex
\documentclass[10pt]{article} 
\usepackage[preprint]{tmlr}

\input{math_commands.tex}

\usepackage{hyperref}
\usepackage{url}
\usepackage{hyperref}
\usepackage{url}
\usepackage{booktabs}       
\usepackage{amsfonts}       
\usepackage{nicefrac}       
\usepackage{microtype}      
\usepackage{xcolor}         
\usepackage{amsmath}
\usepackage{amssymb}
\usepackage{mathtools}
\usepackage{amsthm}
\usepackage{subcaption}
\usepackage{multirow}
\usepackage{cleveref}
\usepackage{adjustbox}
\usepackage{wrapfig}
\usepackage{placeins}
\Crefname{equation}{Eq.}{Eqs.}

\title{MIDAS: Mosaic Input-Specific \\Differentiable Architecture Search}

\author{\name Konstanty Subbotko \email kostek.subbotko@gmail.com \\
      \addr Independent Research}

\begin{document}

\raggedbottom

\maketitle

\input{sec/0_abstract}
\input{sec/1_introduction}
\input{sec/2_relatedwork}
\input{sec/3_methods}
\input{sec/4_experiments}
\input{sec/5_conclusions}

\pagebreak

\bibliography{main}
\bibliographystyle{tmlr}

\input{sec/6_appendix}

\end{document}

%% file: math_commands.tex
\usepackage{amsmath,amsfonts,bm}

\def\eqref#1{equation~\ref{#1}}

\def\1{\bm{1}}

\def\vk{{\bm{k}}}

\def\vq{{\bm{q}}}

\DeclareMathAlphabet{\mathsfit}{\encodingdefault}{\sfdefault}{m}{sl}
\SetMathAlphabet{\mathsfit}{bold}{\encodingdefault}{\sfdefault}{bx}{n}



%% file: sec/0_abstract.tex
\begin{abstract}
Differentiable Neural Architecture Search (NAS) provides efficient, gradient-based methods for automatically designing neural networks, yet its adoption remains limited in practice. We present MIDAS, an approach that modernizes DARTS by replacing static architecture parameters with dynamic, input-specific parameters computed via self-attention. To improve robustness, MIDAS (i) localizes the architecture selection by computing it separately for each spatial patch of the activation map, and (ii) introduces a parameter-free, topology-aware search space that models the two-edge node topology during search. We evaluate MIDAS on the DARTS, NAS-Bench-201, and RDARTS search spaces. In DARTS, it reaches 97.42\% top-1 on CIFAR-10 and 83.38\% on CIFAR-100. In NAS-Bench-201, it discovers optimal or near-optimal architectures. In RDARTS, it achieves the best reported performance among the compared methods on two of four search spaces on CIFAR-10. We further analyze why MIDAS works, showing that patchwise attention improves discrimination among candidate operations and that the resulting input-specific parameter distributions are class-aware.
\end{abstract}

%% file: sec/1_introduction.tex
\section{Introduction}\label{section:introduction}

Neural architecture search (NAS) automates the design of neural networks by exploring a vast space of candidate architectures. In differentiable NAS, the discrete search space is relaxed into a continuous one and optimized via gradient descent~\citep{darts}. Despite promising results~\citep{fbnet,dcnas}, broader adoption remains limited by instability~\citep{robustdarts,evaluation,darts-}. Meanwhile, self-attention–based transformer models have proved effective across various domains~\citep{attisallyouneed,gpt3,vit,clip}. However, when memory and compute are constrained, NAS remains a compelling approach for discovering efficient, hardware-aware architectures~\citep{fbnet}. At the same time, standard differentiable NAS uses the same architecture parameters for all inputs, even though the usefulness of candidate operations may depend on the current features.

\begin{figure*}[!htb]
\begin{center}
\includegraphics[width=\linewidth]{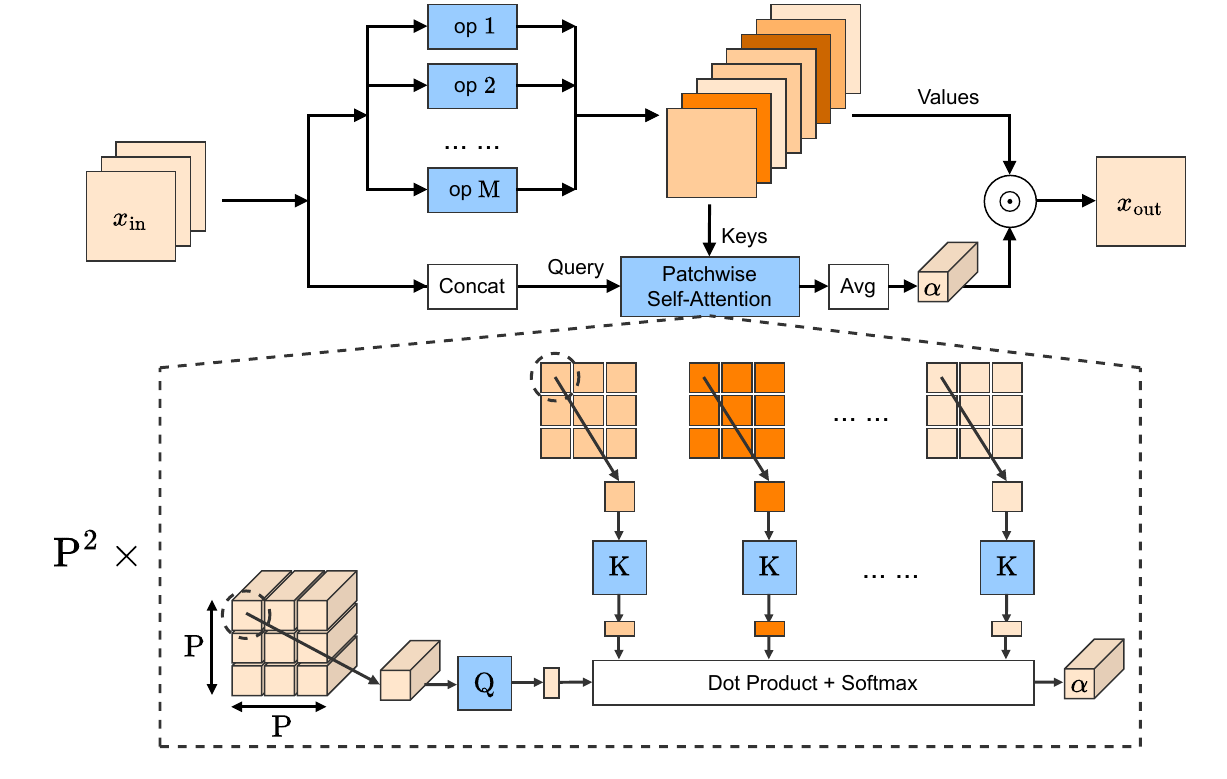}
\caption{Computing input-specific architecture with attention. For a given node, each candidate operation $o^{(j)}$ applied to an incoming feature $x^{(i)}$ produces an activation map $F^{(i,j)} = o^{(j)}(x^{(i)})$. We project the node's concatenated input into a query and the candidate activation maps into keys, and apply dot-product attention to obtain architecture weights. In the mosaic variant, we partition each activation map into $P^2$ patches, compute attention independently within each patch, and average the patch-level distributions to obtain an image-level architecture. For clarity, the topology-aware search space is not illustrated.}
\label{fig:midas}
\end{center}
\end{figure*}

To make architecture selection input-specific, we introduce \textbf{Mosaic Input-specific Differentiable Architecture Search (MIDAS)}, which augments DARTS~\citep{darts} with a lightweight self-attention mechanism that dynamically computes architecture parameters from local features (\Cref{fig:midas}). In DARTS, each node aggregates candidate activation maps using a single set of global learnable architecture parameters as the mixing weights. MIDAS replaces these scalar architecture parameters with input-specific weights computed from the local activations, turning the architecture itself into a distribution of architectures. 

For a given node and input sample, we treat the candidate activation maps $F^{(i,j)} = o^{(j)}(x^{(i)})$ as a set of feature "tokens" and project them into key vectors. We form a single query vector by projecting the node's current input (the concatenation of its incoming features). A dot-product attention between the query and the keys yields a distribution over candidate edges, which we use directly as the local architecture parameters. Intuitively, MIDAS selects architectures that best match the current local context, which allows for more flexibility in architecture selection across the supernet. After convergence, we decode the supernet into a \emph{fixed architecture} by computing a distribution of input-specific parameters across training samples (see \Cref{subsection:decoding}).

While prior work has explored attention \emph{within} the search space as an operation~\citep{attdarts,autoattend}, MIDAS instead uses self-attention to compute architecture parameters as input-dependent functions of features. The most closely related work, AGNAS~\citep{agnas}, computes input-specific gating parameters via global pooling for channel recalibration. MIDAS instead uses dot-product attention between the node input and candidate-operation features.

AGNAS computes gating parameters by first applying global average pooling to the activation maps. However, we observe that this can blur differences among candidate operations due to overly coarse features, particularly in early layers where features are local and spatial dimensions are large. We address this by adopting a \emph{patchwise} design, where all activation maps are divided into $P^2$ spatial patches, and self-attention is applied \emph{independently} within each patch, producing $P^2$ sets of architecture parameters. We call this a \emph{mosaic} approach, since the image-level architecture is composed of patch-level architectures.

Another issue in DARTS concerns capturing cell topology using only operation-wise architecture parameters~\citep{dartspt,dots}, even though the final architecture must select two incoming edges per node. Similar to DOTS~\citep{dots}, we consider all possible pairs of candidate edges. However, unlike DOTS, which introduces a separate topological parameter $\beta_{i_1,i_2}$ alongside standard $\alpha$ parameters, we integrate topology search into our dynamic, self-attention–based mechanism without adding parameters. This formulation models the two-edge node topology during search without introducing additional parameters. The final architecture is obtained by averaging input-specific parameters over a subset of the training data and then selecting both operations and connections based on this average.

MIDAS thus combines self-attention–based, input-specific NAS with local architecture and a parameter-free topology search that accounts for node connectivity. As shown in \Cref{section:experiments}, it achieves competitive results across several search spaces. In the NAS-Bench-201 search space~\citep{nasbench201}, MIDAS consistently finds the optimal or near-optimal architecture; in the DARTS search space~\citep{darts}, it reaches $97.42\%$ accuracy on CIFAR-10; and on RDARTS S1-S4~\citep{robustdarts}, it achieves competitive results across the RDARTS search spaces and sets state of the art on CIFAR-10 for S2 and S4. We also include additional studies that provide deeper insights into the method.

We summarize our main contributions as follows:
\begin{itemize}
    \item \textbf{MIDAS:} using self-attention to compute input-specific architecture parameters, enabling dynamic per-input architecture selection.
    \item \textbf{Mosaic (patchwise) architecture:} local architecture decisions that better discriminate among candidate operations.
    \item \textbf{Parameter-free topology search:} models DARTS-like node connectivity within the same attention mechanism, avoiding extra parameters.
    \item \textbf{Strong results and analysis:} state-of-the-art or competitive performance across multiple search spaces and datasets, with additional ablation studies.
\end{itemize}

%% file: sec/2_relatedwork.tex
\section{Related Work}\label{section:related_work}

\subsection{Preliminaries}\label{subsection:preeliminaries}

\paragraph{DARTS Search Space.}

In DARTS~\citep{darts}, the \emph{supernet} is a sequence of $N$ convolutional cells, each either a normal or a reduction cell~\citep{nasnet}. A cell is a densely-connected directed acyclic graph (DAG) with $N$ nodes.

Let $\mathcal{O}=\{o^{(1)}, \ldots, o^{(M)}\}$ be the set of $M$ candidate operations. Consider the $k$-th node. For each possible input $x^{(i)} \in \mathbb{R}^{B \times C \times H \times W}$ and operation $o^{(j)} \in \mathcal{O}$, 
DARTS assigns an edge-specific learnable parameter \(\alpha_k^{(i,j)}\) that indicates the relative importance of applying \(o^{(j)}\) to \(x^{(i)}\).
Architecture parameters are shared across cells of the same type. For readability, we omit the node index \(k\) below.
\begin{equation}
    F^{(i,j)} = o^{(j)}(x^{(i)}), \quad F^{(i,j)} \in \mathbb{R}^{B \times C \times H \times W}.
\end{equation}
The output of the $k$-th node is a weighted sum over all such candidate activation maps:
\begin{equation}\label{eq:mixops}
    x_{\text{out}} = \sum_{i=1}^{k+1}\sum_{j=1}^{M}p^{(i,j)}F^{(i,j)},
\end{equation}
where
\begin{equation}\label{eq:scalars}
    p^{(i,j)}= \frac{\exp(\alpha^{(i,j)})}{\sum_{j'=1}^{M} \exp(\alpha^{(i,j')})}.
\end{equation}
The cell output is obtained by concatenating the outputs of all intermediate nodes.

\paragraph{Optimization.}
Because the supernet is differentiable, convolution weights $\omega$ and architecture parameters $\alpha$ are optimized jointly via a bilevel objective:
\begin{equation}
    \min_{\alpha} \, \mathcal{L}_{\text{A}}(\omega^*, \alpha)
    \quad \text{s.t.} \quad
    \omega^* = \arg\min_{\omega} \, \mathcal{L}_{\text{B}}(\omega, \alpha),
\end{equation}
where $\mathcal{L}_{\text{A}}$ and $\mathcal{L}_{\text{B}}$ are cross-entropy losses on disjoint splits of the training data A and B. Solving this exactly is expensive, so DARTS relies on an alternating approximation:
\begin{itemize}
    \item Update architecture parameters $\alpha$ using $\nabla_{\alpha} \mathcal{L}_{\text{A}}(\omega, \alpha)$,
    \item Update convolution weights $\omega$ using $\nabla_{\omega} \mathcal{L}_{\text{B}}(\omega, \alpha)$.
\end{itemize}
After convergence, DARTS selects the two strongest incoming edges for each node and retains the strongest non-zero operation on each selected edge.

\subsection{Rank disorder in DARTS}\label{subsection:dartsissues}

Although DARTS~\citep{darts} was pivotal in developing differentiable NAS, later work~\citep{robustdarts,darts-,evaluatingsearch,dots} revealed several limitations. One of the major shortcomings is \emph{rank disorder}, where operation weights fail to reflect operation's true importance~\citep{evaluatingsearch}.
DARTS’s decoding procedure can exacerbate rank disorder, as selecting the top two incoming edges based on operation-wise architecture parameters may be misleading. P-DARTS~\citep{progressivedarts} partially mitigates it by progressively pruning weaker operations, and DARTS-PT~\citep{dartspt} keeps only operations whose removal causes the largest decrease in accuracy. DOTS~\citep{dots} observes that edge-level importance is poorly captured by operation-wise parameters and proposes a topology search that assigns distinct parameters to every pair of candidate edges.

\subsection{Input-Specific NAS}

Attention mechanisms emerged in 2014 for RNN-based image classification~\citep{firstatt} and have since been applied for various purposes, including channel recalibration and input-specific aggregation~\citep{nicg,sea,cbam}. The introduction of the self-attention Transformer~\citep{attisallyouneed} revolutionized sequence modeling, inspiring extensive research on attention-based architectures.

NAS has leveraged attention in several ways. For instance, ADARTS~\citep{adarts} employs attention to identify the most relevant channels during the search, improving efficiency of the supernet. Other methods directly search for attention modules~\citep{autoattend,dartformer}, optimizing their structure for a given task. InstaNAS~\citep{instanas} is an example of input-specific NAS, as it uses a controller that dynamically selects an architecture for each data sample. Meanwhile, AGNAS~\citep{agnas} instead utilizes channel-wise attention weights from a shallow MLP to recalibrate concatenated candidate activations before aggregation.
These channel-wise attention weights can be viewed as measures of operation importance.

%% file: sec/3_methods.tex
\section{Methods}\label{section:methods}

In this section, we describe how MIDAS computes input-specific architecture parameters using a lightweight dot-product attention module (\Cref{fig:midas}). We use the same notation as in \Cref{subsection:preeliminaries} for consistency.

\subsection{Input-Specific Architecture}

Motivated by the success of attention-based models~\citep{attisallyouneed}, we replace static architecture parameters with dynamic parameters computed via patchwise self-attention. This increases expressiveness by adapting the architecture to each input.

\paragraph{Local Architecture.}

Consider the $k$-th node. Let $F^{(i,j)} \in \mathbb{R}^{B \times C \times H \times W}$ be the activation map obtained by applying the $j$-th operation $o^{(j)}$ to the $i$-th input $x^{(i)}$. Let $P$ denote the number of patches along each spatial dimension, so that each activation map is partitioned into $P^2$ non-overlapping local patches:
\begin{equation}
    F_{u, v}^{(i,j)} \in \mathbb{R}^{B \times C \times \tfrac{H}{P} \times \tfrac{W}{P}}
    \quad\text{for}\hspace{2mm} 1 \leq u, v \leq P.
\end{equation}
We then apply global average pooling within each patch:
\begin{equation}\label{eq:patches}
\widetilde{F}_{u,v}^{(i,j)}\hspace{2mm}=\hspace{2mm}\frac{P^2}{H \times W}
\sum_{p=1}^{\tfrac{H}{P}} 
\sum_{q=1}^{\tfrac{W}{P}}
F_{u,v}^{(i,j)}(p, q),
\end{equation}
yielding $\widetilde{F}_{u,v}^{(i,j)} \in \mathbb{R}^{B \times C \times 1 \times 1}$. This patchwise design lets local features contribute differently to the architecture decisions. We refer to this as \emph{mosaic}, since the image-level architecture is composed of patch-level architectures. In contrast, a single global pooling over the entire activation map can produce overly coarse features, especially in earlier layers where representations are local and spatial resolution is large.

\paragraph{Computing Keys and Query.}

We learn \emph{keys} from the pooled features using a shallow two-layer MLP:
\begin{equation}
    \vk_{u,v}^{(i,j)} = K\bigl(\widetilde{F}_{u,v}^{(i,j)}\bigr) \;\in\; \mathbb{R}^{B \times C},
\end{equation}
where
\begin{equation}
\begin{aligned}
    &K(x) = W_{k_2} \,\mathrm{LeakyReLU}\bigl(W_{k_1}x\bigr),\\
    &W_{k_1}, W_{k_2} \;\in\; \mathbb{R}^{C \times C}.
\end{aligned}
\end{equation}
Similarly, to obtain a \emph{query}, we concatenate the $(k+1)$ node inputs $\big[x^{(1)}, x^{(2)}, \ldots, x^{(k+1)}\big]$ along the channel dimension:
\begin{equation}
    X \;\in\; \mathbb{R}^{B \times \bigl(C \cdot (k+1)\bigr) \times H \times W}.
\end{equation}
We again partition $X$ into $P^2$ patches and apply global average pooling to form \linebreak $\widetilde{X}_{u,v} \in \mathbb{R}^{B \times \bigl(C (k+1)\bigr) \times 1 \times 1}$, then map each pooled patch to a query vector:
\begin{equation}
    \vq_{u,v} = Q\bigl(\widetilde{X}_{u,v}\bigr) \;\in\; \mathbb{R}^{B \times C},
\end{equation}
where
\begin{equation}
\begin{aligned}
    &Q(x) = W_{q_2}\,\mathrm{LeakyReLU}\bigl(W_{q_1}x\bigr),\\
    &W_{q_1} \;\in\; \mathbb{R}^{C \times \bigl(C\cdot(k+1)\bigr)}, 
    \quad W_{q_2} \;\in\; \mathbb{R}^{C \times C}.
\end{aligned}
\end{equation}

\paragraph{Topology-Aware Search Space.}
Because DARTS selects two incoming edges per node, we define a topology search space over unordered pairs of candidate edges. For node \(k\), let:
\begin{equation}
\mathcal P_k=
\left\{
\{(i_1,j_1),(i_2,j_2)\}:
1\le i_1<i_2\le k+1,\;
1\le j_1,j_2\le M
\right\}.  
\end{equation}
The ordering ensures that each valid pair appears once and that the two edges originate from different inputs. For patch \((u,v)\), we compute an unnormalized attention score:
\begin{equation}
    \alpha_{u,v}^{(i_1,j_1,i_2,j_2)}
    = \frac{\bigl(\vk_{u,v}^{(i_1,j_1)} + \vk_{u,v}^{(i_2,j_2)}\bigr) 
    \,\cdot\,\vq_{u,v}}{\sqrt{C}},
\end{equation}
which scores the joint selection of operations \(o^{(j_1)}\) and \(o^{(j_2)}\) on inputs \(x^{(i_1)}\) and \(x^{(i_2)}\), respectively, within patch \((u,v)\). We normalize these scores across all valid pairs using softmax:
\begin{equation}
p_{u,v}^{(i_1,j_1,i_2,j_2)}
=
\frac{
\exp\left(\alpha_{u,v}^{(i_1,j_1,i_2,j_2)}\right)
}{
\sum_{\{(i'_1,j'_1),(i'_2,j'_2)\}\in\mathcal{P}_k}
\exp\left(\alpha_{u,v}^{(i'_1,j'_1,i'_2,j'_2)}\right)
}.
\end{equation}
For stability, we average the patch-level distributions to obtain a single image-level pair distribution:
\begin{equation}
    p^{(i_1,j_1,i_2,j_2)}
    = \frac{1}{P^2}\sum_{u=1}^{P}\sum_{v=1}^{P} 
      p_{u,v}^{(i_1,j_1,i_2,j_2)}.\label{eq:avg}
\end{equation}

Let $\boldsymbol{\alpha}_{u,v}$ denote the vector of attention logits over all valid candidate pairs for patch $(u,v)$. In general,
\begin{equation}
\frac{1}{P^2}\sum_{u,v}\mathrm{softmax}\!\left(\boldsymbol{\alpha}_{u,v}\right)
\neq
\mathrm{softmax}\!\left(
\frac{1}{P^2}\sum_{u,v}\boldsymbol{\alpha}_{u,v}
\right).
\end{equation}
Thus, averaging patch-level distributions preserves information that would be lost by averaging the logits first. Empirically, we observe that the proposed patchwise attention improves discrimination among candidate operations in early cells (\Cref{subsection:patch}).

We then obtain the marginal importance of an individual edge by summing the probabilities of all valid pairs that contain it:
\begin{equation}
p^{(i,j)}
=
\sum_{\substack{\pi\in\mathcal{P}_k\\(i,j)\in\pi}}
p^{\pi}.\label{eq:marginal}
\end{equation}
Since every pair contributes to the marginals of both constituent edges, the total marginal probability mass is
\begin{equation}
\sum_{i=1}^{k+1}\sum_{j=1}^{M}p^{(i,j)}
=
2.
\end{equation}
These marginals are used as edge weights during search and also form the basis for final decoding. The node output is therefore
\begin{equation}
    x_{\text{out}} = \sum_{i, j} p^{(i, j)} F^{(i,j)}.
\end{equation}
Despite the large combination space, the computation can be implemented efficiently using tensor operations rather than explicit iteration. We quantify the resulting search-phase overhead in \Cref{subsection:darts}.

\paragraph{Complexity.}

MIDAS applies attention over the small set of candidate (input, operation) choices at a node, not over all spatial positions. The token count is bounded by (\#node inputs $\times$ \#operations). For a fixed patch size, increasing spatial resolution increases the number of patches, and the attention cost scales linearly with this number.

\subsection{Searching Protocol}\label{subsection:decoding}

Following DARTS, we adopt an approximate bilevel optimization scheme on two disjoint splits of the training data: one split updates the network weights $\omega$ and the other updates the architecture parameters. In MIDAS, the latter include the attention projection networks $\{W_{k_1}, W_{k_2}, W_{q_1}, W_{q_2}\}$ for each node. These parameters are \emph{not} shared across nodes.

\paragraph{Decoding.}
Because MIDAS is input-specific, it does not yield a single fixed set of architecture parameters. After convergence, for each sample in the decoding subset we first average the patch-level pair distributions according to \Cref{eq:avg} and compute the marginals \(p^{(i,j)}\) using \Cref{eq:marginal}. We then average these marginals across samples. We denote the resulting mean marginal by \(\bar p^{(i,j)}\). 
The top-two candidate edges from different inputs per node are selected according to \(\bar p^{(i,j)}\). We therefore decode from the averaged marginals rather than from the highest-probability pair, since they better reflect operation strength. Additional details and an ablation on the decoding subset are provided in \Cref{app:decoding}.

%% file: sec/4_experiments.tex
\begin{table*}[t]
\caption{Results on NAS-Bench-201~\citep{nasbench201}. We report accuracy of the found architectures on CIFAR-10, CIFAR-100, and ImageNet-16-120.}
\label{table:nasbench}
\begin{center}
\begin{adjustbox}{width=\textwidth}
\begin{small}
\begin{sc}
\small
\begin{tabular}{lcccccc}
\toprule
\multirow{2}{*}{\textbf{Methods}} & \multicolumn{2}{c}{\textbf{CIFAR-10}} & \multicolumn{2}{c}{\textbf{CIFAR-100}} & \multicolumn{2}{c}{\textbf{ImageNet16-120}} \\
\cmidrule(lr){2-3} \cmidrule(lr){4-5} \cmidrule(lr){6-7}
& \textbf{valid} & \textbf{test} & \textbf{valid} & \textbf{test} & \textbf{valid} & \textbf{test} \\
\midrule
DARTS~\citep{darts} & 39.77$\pm$0.00 & 54.30$\pm$0.00 & 15.03$\pm$0.00 & 15.61$\pm$0.00 & 16.43$\pm$0.00 & 16.32$\pm$0.00 \\
GDAS~\citep{gdas} & 89.89$\pm$0.08 & 93.61$\pm$0.09 & 71.34$\pm$0.04 & 70.70$\pm$0.30 & 41.59$\pm$1.33 & 41.71$\pm$0.98 \\
PC-DARTS~\citep{pcdarts} & 89.96$\pm$0.15 & 93.41$\pm$0.30 & 67.12$\pm$0.39 & 67.48$\pm$0.89 & 40.83$\pm$0.08 & 41.31$\pm$0.22 \\
iDARTS~\citep{idarts} & 89.86$\pm$0.60 & 93.58$\pm$0.32 & 70.57$\pm$0.24 & 70.83$\pm$0.48 & 40.38$\pm$0.59 & 40.89$\pm$0.68 \\
DARTS-~\citep{darts-} & 91.03$\pm$0.44 & 93.80$\pm$0.40 & 71.36$\pm$1.51 & 71.53$\pm$1.51 & 44.87$\pm$1.46 & 45.12$\pm$0.82 \\
AGNAS~\citep{agnas} & 91.25$\pm$0.02 & 94.05$\pm$0.06 & 72.4$\pm$0.38 & 72.41$\pm$0.06 & 45.50$\pm$0.00 & 45.98$\pm$0.46 \\
$\beta$-DARTS~\citep{betadarts} & \textbf{91.55$\pm$0.00} & \textbf{94.36$\pm$0.00} & \textbf{73.49$\pm$0.00} & \textbf{73.51$\pm$0.00} & \textbf{46.37$\pm$0.00} & \textbf{46.34$\pm$0.00} \\
\midrule
Ours (C10, best) & \textbf{91.55} & \textbf{94.36} & \textbf{73.49} & \textbf{73.51} & \textbf{46.37} & \textbf{46.34} \\
Ours (C10) & 91.21$\pm$0.68 & 94.21$\pm$0.30 & 72.80$\pm$1.39 & 72.91$\pm$1.20 & 44.97$\pm$2.80 & 45.12$\pm$2.45 \\
Ours (C100) & 90.20$\pm$1.93 & 93.00$\pm$1.57 & 70.34$\pm$2.13 & 70.58$\pm$2.15 & 44.08$\pm$2.54 & 43.29$\pm$2.45 \\
\midrule
optimal   & 91.61 & 94.37 & 73.49 & 73.51 & 46.77 & 47.31 \\
\bottomrule
\end{tabular}
\end{sc}
\end{small}
\end{adjustbox}
\end{center}
\end{table*}

\begin{table*}[t]
\caption{Results on CIFAR-10 and CIFAR-100 in the DARTS search space. We report architectures discovered on CIFAR-10 and CIFAR-100. For MIDAS, we report mean$\pm$std over five independent retraining runs. An asterisk (*) indicates different architectures for CIFAR-10 and CIFAR-100.}

\label{table:cifar}
\begin{center}
\begin{adjustbox}{width=\textwidth}

\begin{small}
\begin{sc}
\begin{tabular}{lcccc}
\toprule
\textbf{Methods} & \textbf{GPU (days)} & \textbf{Params (M)} & \textbf{CIFAR-10} & \textbf{CIFAR-100}\\
\midrule
NASNet-A~\citep{nasnet} & 2000 & 3.3 & 97.35 & 83.18 \\
SNAS~\citep{snas} & 1.5 & 2.8 & 97.15 $\pm$ 0.02 & 82.45 \\
DARTS (1st)~\citep{darts} & 0.4 & 3.4 & 97.00 $\pm$ 0.14 & 82.46 \\
DARTS (2nd)~\citep{darts} & 1 & 3.3 & 97.24 $\pm$ 0.09 & $-$ \\
P-DARTS~\citep{pdarts} & 0.3 & 3.3 & 97.19 $\pm$ 0.14 & $-$ \\
PC-DARTS~\citep{pcdarts} & 0.1 & 3.6 & 97.43 $\pm$ 0.07 & $-$ \\
SDARTS-ADV~\citep{sdarts} & 1.3 & 3.3 & 97.39 $\pm$ 0.02 & $-$ \\
DOTS~\citep{dots} & 0.3 & 3.5 / 4.1* & \textbf{97.51 $\pm$ 0.06} & \textbf{83.52 $\pm$ 0.13} \\
DARTS+PT~\citep{dartspt} & 0.8 & 3.0 & 97.39 $\pm$ 0.08 & $-$ \\
DARTS-~\citep{darts-} & 0.4 & 3.5 / 3.4* & 97.41 $\pm$ 0.08 & 82.49 $\pm$ 0.25 \\
AGNAS~\citep{agnas} & 0.4 & 3.6 & 97.47 $\pm$ 0.00 & $-$ \\
$\beta$-DARTS~\citep{betadarts} & 0.4 & 3.8 & 97.49 $\pm$ 0.07 & 83.48 $\pm$ 0.03 \\
\midrule
Ours (C10) & 0.5 & 3.5 & 97.42 $\pm$ 0.06 & 83.38 $\pm$ 0.40 \\
Ours (C100) & 0.5 & 3.3 & 97.25 $\pm$ 0.09 & 82.67 $\pm$ 0.20 \\
\bottomrule
\end{tabular}
\end{sc}
\end{small}
\end{adjustbox}
\end{center}
\end{table*}

\section{Experiments}\label{section:experiments}
We evaluate MIDAS across multiple search spaces (NAS-Bench-201~\citep{nasbench201}, DARTS~\citep{darts}, and S1-S4~\citep{robustdarts}) and three datasets (CIFAR-10, CIFAR-100~\citep{cifar}, and ImageNet~\citep{imagenet}) to assess generalizability.

\begin{table*}[t]
\caption{ImageNet results when transferring cells discovered on CIFAR-10.}
\label{table:imagenet}
\begin{center}
\begin{small}
\begin{sc}
\begin{tabular}{lcccc}
\toprule
\textbf{Methods} & \textbf{Params (M)} & \textbf{Top1 (\%)} & \textbf{Top5 (\%)}\\
\midrule
NASNet-A~\citep{nasnet} & 5.3 & 74.0 & 91.6 \\
SNAS~\citep{snas} & 4.3 & 72.7 & 90.8 \\
DARTS~\citep{darts} & 4.7 & 73.3 & 91.3 \\
P-DARTS~\citep{pdarts} & 5.1 & 75.3 & 92.5 \\
SDARTS-ADV~\citep{sdarts} & 5.4 & 74.8 & 92.2 \\
DOTS~\citep{dots} & 5.2 & 75.7 & 92.6 \\
DARTS+PT~\citep{dartspt} & 4.6 & 75.5 & 92.0 \\
$\beta$-DARTS~\citep{betadarts} & 5.5 & \textbf{76.1} & \textbf{93.0}  \\
\midrule
Ours & 5.2 & 75.4 & 92.5 \\
\bottomrule
\end{tabular}
\end{sc}
\end{small}
\end{center}
\end{table*}

\begin{table*}[t]
\caption{Search-phase overhead of MIDAS relative to DARTS in the DARTS search space. We report parameter count, FLOPs, peak GPU memory, and total search time (measured on a V100 GPU).}
\label{table:overhead}
\begin{center}
\begin{adjustbox}{width=\textwidth}
\begin{small}
\begin{sc}
\begin{tabular}{ccccc}
\toprule
\textbf{Method} & \textbf{Params (M)} & \textbf{FLOPs (M)} & \textbf{Peak GPU mem (MB)} & \textbf{Search time (GPU h)}\\
\midrule
DARTS & 2.02 & 659 & 6481 & 4.2 \\
\midrule
MIDAS & 2.43 & 673 & 8447 & 5.2 \\
\bottomrule
\end{tabular}
\end{sc}
\end{small}
\end{adjustbox}
\end{center}
\end{table*}

\begin{table*}[t]
\caption{Results on S1–S4. Due to the topology-aware search space, S2 and S3 are equivalent for MIDAS.}
\label{table:rdarts}
\begin{center}
\begin{adjustbox}{width=\textwidth}
\begin{small}
\begin{sc}
\begin{tabular}{ccccccccc}
\toprule
\textbf{Dataset} & \textbf{Space} & \textbf{DARTS} & \textbf{RDARTS} & \textbf{DARTS-}& \textbf{DARTS+PT} & \textbf{Shapley-NAS} & \textbf{RF-DARTS} & \textbf{Ours} \\
\midrule
\multirow{4}{*}{CIFAR-10}
& S1 & 3.84 & 2.78 & \textbf{2.68} & 3.50 & 2.82 & 2.95 & 2.81 \\
& S2 & 4.85 & 3.31 & 2.63 & 2.79 & 2.55 & 4.21 & \textbf{2.49} \\
& S3 & 3.34 & 2.51 & \textbf{2.42} & 2.49 & \textbf{2.42} & 2.83 & 2.49 \\
& S4 & 7.20 & 3.56 & 2.86 & 2.64 & 2.63 & 3.33 & \textbf{2.59} \\
\midrule
\multirow{4}{*}{CIFAR-100}
& S1 & 29.46 & 25.93 & \textbf{22.41} & 24.48 & 23.60 & 22.75 & 23.50 \\
& S2 & 26.05 & 22.30 & \textbf{21.61} & 23.16 & 22.77 & 22.18 & 22.18 \\
& S3 & 28.90 & 22.36 & \textbf{21.13} & 22.03 & 21.92 & 24.67 & 22.18 \\
& S4 & 22.85 & 22.18 & \textbf{21.55} & 20.80 & 21.53 & 21.19 & 20.95 \\
\bottomrule
\end{tabular}
\end{sc}
\end{small}
\end{adjustbox}
\end{center}
\end{table*}
\pagebreak
\subsection{Experiment Details}\label{subsection:expdetails}

\paragraph{Patch Selection.}

We partition each activation map into non-overlapping patches of size $PS \times PS$. For a feature map of height $H$ and width $W$, this yields $P = H/PS = W/PS$ patches (assuming divisibility). If $PS$ is too large, it makes operations harder to discriminate, and an overly small $PS$ can downweight large–receptive-field operations. In the DARTS space we set $PS=8$ to accommodate larger-context operations such as dilated convolutions. For NAS-Bench-201, which uses only $3\times3$ convolutions, we set $PS=4$. We provide our rationale in \Cref{app:ps}.

\paragraph{Following Best NAS Practices.}

To ensure quality and reproducibility, we adhere to recommended NAS practices~\citep{bestpractices} and release code for both search and retraining. Aside from a few principled choices, we avoid extensive hyperparameter tuning to reduce benchmark overfitting. All experiments use four fixed seeds \(\{1,2,3,4\}\). For DARTS, the supernet with the best validation accuracy is selected before retraining, as described in \Cref{subsection:darts}. Full results and protocols are reported in \Cref{app:expdetails,app:protocols}.

\subsection{NAS-Bench-201 Benchmark}

NAS-Bench-201 defines a reduced DARTS-like space with four nodes and five operations per edge (15{,}625 architectures), along with precomputed ground-truth performance on CIFAR-10, CIFAR-100, and ImageNet16-120. Following the standard evaluation protocol, we run four searches on CIFAR-10 and report mean$\pm$std across the four search runs using the benchmark’s provided tabular accuracies.

Because NAS-Bench-201 uses a single shared cell, during decoding we average input-specific parameters across cells to obtain one normal and one reduction cell. Results are shown in \Cref{table:nasbench}.
Across the four CIFAR-10 searches, MIDAS finds optimal or near-optimal architectures, reaching the best result in three runs (see \Cref{app:expdetails}).
Although this is uncommon, we also search directly on CIFAR-100 to test robustness. As expected, variance increases with task difficulty, but MIDAS still discovers strong architectures on average.

\subsection{DARTS Search Space}\label{subsection:darts}

We next evaluate MIDAS in the DARTS search space~\citep{darts}. DARTS uses shared normal and reduction cells during both search and retraining. Because MIDAS computes different architectures in each node, to decode MIDAS's input-specific parameters into fixed cells, we adopt the AGNAS~\citep{agnas} decoding procedure, which decodes one cell per spatial level, yielding three normal and two reduction cells. See \Cref{app:decoding} for details and visualizations.

We follow the standard evaluation protocol: conduct four searches, select the supernet with the best validation accuracy, and retrain the derived architecture from scratch for 600 epochs on CIFAR-10 and CIFAR-100. As shown in \Cref{table:cifar}, MIDAS is competitive with prior work, reaching $97.42\pm0.06\%$ on CIFAR-10 and $83.38\pm0.40\%$ on CIFAR-100. Searching directly on CIFAR-100 yields comparable but slightly lower accuracy. To assess transferability, we train the CIFAR-10–discovered cell on ImageNet~\citep{imagenet}. \Cref{table:imagenet} again shows competitive results.

\paragraph{Overhead vs. DARTS.}

Since MIDAS computes input-specific architecture parameters, it incurs additional cost over DARTS. We quantify this overhead in the DARTS search space in \Cref{table:overhead}. MIDAS increases parameters by approximately $0.41$M and FLOPs by $2\%$. Peak memory increases from $6481$ to $8447$ MB, and search time from $4.2$ to $5.2$ GPU-hours on a V100. The additional cost comes from the shallow attention MLPs and patchwise attention, and is limited to the search phase.

\subsection{Robustness of MIDAS}\label{subsection:robustness}

~\citet{robustdarts} propose four search spaces S1-S4 to evaluate the robustness of DARTS. We report results on CIFAR-10 and CIFAR-100 in \Cref{table:rdarts}. Following the RDARTS evaluation protocol, we retrain the best architecture with 20 layers/36 channels on CIFAR-10 and 8 layers/16 channels on CIFAR-100. Since S3 is identical to S2 except for the addition of the \textit{Zero} operation, it is equivalent under MIDAS's topology-aware search space, which does not require \textit{Zero}. 
MIDAS remains competitive across all four search spaces. In particular, on S2 and S4 it achieves the best CIFAR-10 results among the compared methods, with $2.49\%$ and $2.59\%$ test error, respectively.

\subsection{Ablation Studies}\label{subsection:patch}

\begin{figure}
\begin{center}
\includegraphics[width=\textwidth]{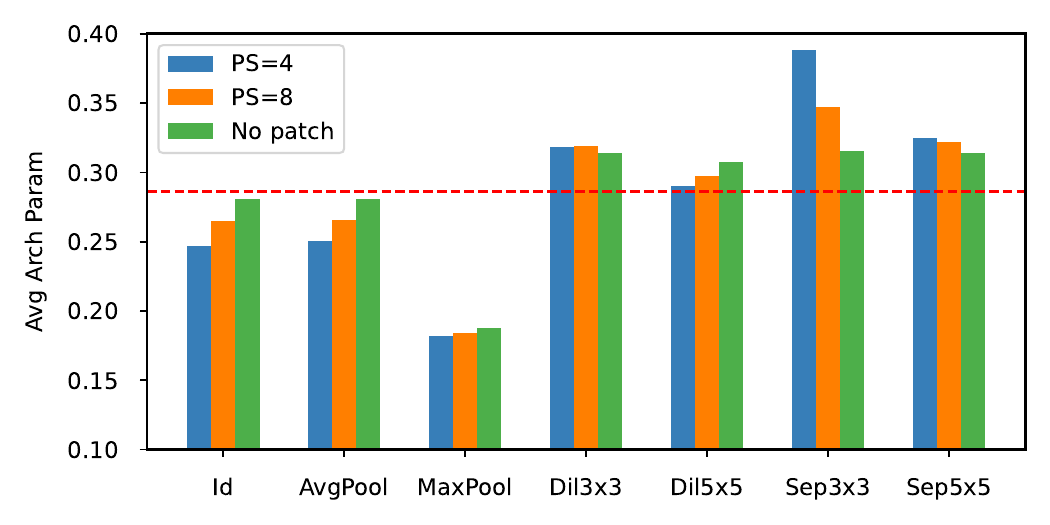}
\caption{Learned input-specific architecture parameters in the first two cells in the DARTS search space on CIFAR-10, averaged over four runs. We compare three variants: \textit{no patch} (global average pooling only), \textit{PS=4} (patch size $4\times4$), and \textit{PS=8} (patch size $8\times8$). The horizontal line denotes uniform importance across operations. We observe that \textit{no patch} fails to discriminate among learnable operations, essentially assigning similar weights to all four learnable operations.}
\label{fig:patch_abl}
\end{center}
\end{figure}

\paragraph{Patchwise Attention.}

We test whether partitioning into patches improves discrimination among candidate operations. Namely, we compare three variants on CIFAR-10: (i) no partition (global average pooling only), (ii) $PS=4$, and (iii) $PS=8$. \Cref{fig:patch_abl} shows the average input-specific parameter for each operation in the first two cells.
With \(PS=4\) or \(PS=8\), MIDAS clearly differentiates among learnable operations in the early cells, whereas global pooling assigns near-uniform weights due to overly coarse global features.

\paragraph{Search Hyperparameters and Attention.}

\begin{table*}[t]
\caption{Ablations on NAS-Bench-201. DARTS (MIDAS hparams) uses the MIDAS search hyperparameters. The learned-query variant replaces the input-specific query with a learned query for each block. The shared-key variant uses one key projection across blocks. Results are mean $\pm$ standard deviation over four search runs.}
\label{table:ablation}
\begin{center}
\begin{adjustbox}{width=\textwidth}
\begin{small}
\begin{sc}
\small
\begin{tabular}{lcccccc}
\toprule
\multirow{2}{*}{\textbf{Methods}} & \textbf{CIFAR-10} & \multicolumn{2}{c}{\textbf{CIFAR-100}} & \multicolumn{2}{c}{\textbf{ImageNet16-120}} \\
\cmidrule(lr){2-2} \cmidrule(lr){3-4} \cmidrule(lr){5-6}
 & \textbf{test} & \textbf{valid} & \textbf{test} & \textbf{valid} & \textbf{test} \\
\midrule
DARTS (MIDAS hparams) & 88.95$\pm$2.40 & 61.58$\pm$3.84 & 62.08$\pm$3.91 & 31.30$\pm$6.16 & 30.69$\pm$6.28 \\
MIDAS, learned query & 92.52$\pm$1.34 & 67.72$\pm$2.82 & 67.89$\pm$2.82 & 35.00$\pm$7.35 & 34.91$\pm$7.14 \\
MIDAS, shared key & 93.61$\pm$1.07 & 70.79$\pm$3.40 & 71.10$\pm$2.98 & 42.42$\pm$6.37 & 42.34$\pm$6.22 \\
MIDAS & 94.21$\pm$0.30 & 72.80$\pm$1.39 & 72.91$\pm$1.20 & 44.97$\pm$2.80 & 45.12$\pm$2.45 \\
\bottomrule
\end{tabular}
\end{sc}
\end{small}
\end{adjustbox}
\end{center}
\end{table*}

\begin{figure}
    \centering
    \begin{subfigure}[t]{0.49\linewidth}
        \centering
        \includegraphics[width=\linewidth]{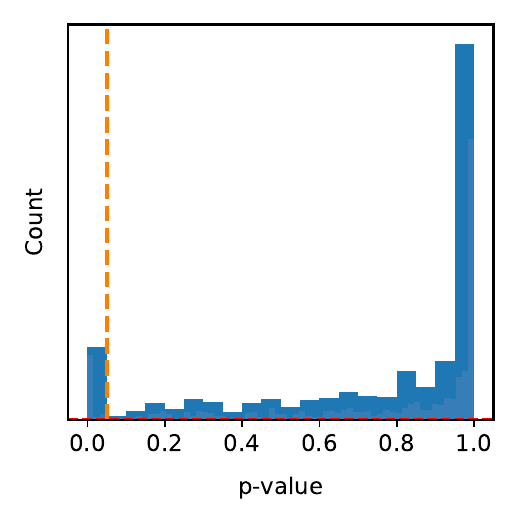}
        \label{fig:unimodal}
    \end{subfigure}
    \hfill
    \begin{subfigure}[t]{0.49\linewidth}
        \centering
        \includegraphics[width=\linewidth]{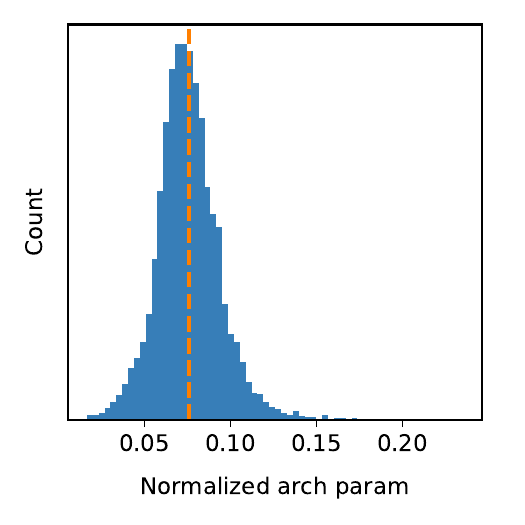}
        \label{fig:distribution}
    \end{subfigure}
    \caption{(Left) Histogram of $p$-values from Hartigan's dip test across all architecture parameters. The vertical line marks the $0.05$ rejection threshold. Most $p$-values exceed $0.05$, so unimodality is not rejected for the majority. (Right) Distribution of an example input-specific parameter, with the vertical line marking the mean value.}
    \label{fig:distribution}
\end{figure}

We next isolate the effect of the MIDAS parameterization from the modified search settings and examine two design choices in the attention module. The results are reported in \Cref{table:ablation}.
Applying MIDAS search hyperparameters to DARTS improves its performance, but does not recover MIDAS performance. This suggests that the gains are not explained by the search hyperparameters alone. Replacing the input-specific query with a learned query also reduces performance, supporting the use of input-dependent queries. The shared-key variant performs closer to MIDAS, but remains consistently lower. These variants use the hyperparameters selected for MIDAS.

\subsection{Analysis of Input-Specific Architecture}

In MIDAS, each architecture parameter is modeled as a distribution over samples rather than a single scalar. As shown in \Cref{fig:variance}, the parameters vary meaningfully across inputs, with variance increasing toward the end of training instead of collapsing into a sharply peaked distribution.

\begin{figure}
    \centering
    \includegraphics[width=0.85\linewidth]{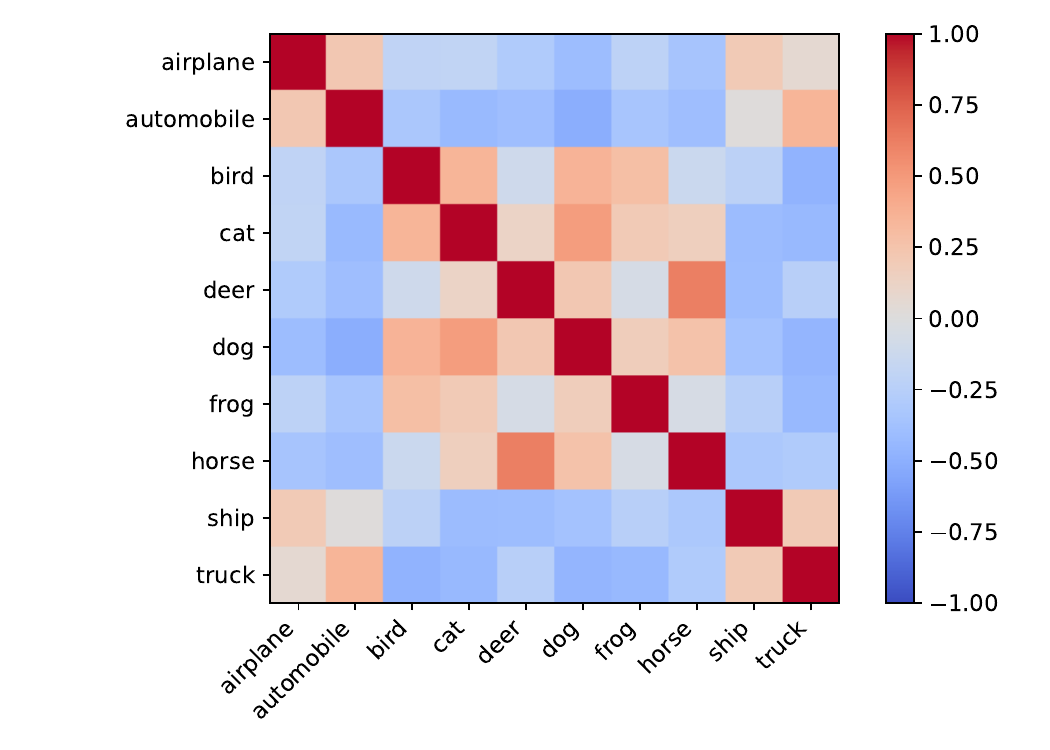}
    \caption{Cosine similarity between CIFAR-10 classes derived from input-specific architecture parameters (higher is more similar). Best viewed in color.}
    \label{fig:class-sim}
\end{figure}

\paragraph{Unimodality.}
Ultimately, we must decode one architecture for the entire data distribution. A potential pitfall for MIDAS is \emph{multimodality}, where different samples could produce conflicting architecture decisions, making operation importance hard to quantify. In an extreme case, if a parameter is near $0$ for half the samples and near $1$ for the other half, decoding becomes ambiguous. We therefore test for multimodality using Hartigan's dip test~\citep{hartiganDip} on all distributions of architecture parameters (computed over $5000$ samples). In \Cref{fig:distribution} (left), we plot the histogram of $p$-values with a $0.05$ rejection threshold. Most \(p\)-values exceed $0.05$, and unimodality is not rejected for $91\%$ of parameters. This does not establish unimodality, but the test does not find evidence against it for most parameters. \Cref{fig:distribution} (right) shows an example distribution for a randomly selected operation.

To test whether pooling classes affects this result, we repeat the analysis separately within each CIFAR-10 class. For $95\%$ of parameters, none of the class-conditional distributions rejects unimodality. 
Additionally, for each architecture parameter, we compute its median over samples separately for each CIFAR-10 class, then take the range of these ten class medians and normalize it by the parameter standard deviation. A value of 1 means that the most separated class medians differ by one standard deviation of that parameter, with larger values indicating stronger class-dependent shifts. The median value is $1.28$ across all parameters, with values of $0.90$, $1.31$, and $1.39$ for early, middle, and late cells, respectively. This indicates stronger class-dependent variation in later cells.

\paragraph{Class-Aware Architecture.}

To test whether these architectural variations reflect noise or meaningful structure, we plot cosine similarities between classes derived from input-specific architecture parameters in \Cref{fig:class-sim}. Namely, we take a converged CIFAR-10 supernet and compute input-specific architecture parameters for 5000 images (500 per class). For each class, we average the parameters over its 500 samples and restrict the analysis to the last two lowest-resolution cells, resulting in a 196-dimensional vector per class (since there are 98 architecture parameters per cell in the DARTS search space). We then z-score each parameter across classes to standardize the scales and compute pairwise cosine similarities, producing the $10 \times 10$ similarity matrix shown in \Cref{fig:class-sim}.

We observe that the class similarities qualitatively align with semantic relationships, suggesting that MIDAS captures class-related structure rather than random, seed-dependent variation. For instance, architectures obtained from vehicle classes (airplane, automobile, ship, truck) are highly similar to each other, while animal classes form a separate cluster (with horse and deer being particularly close). This is consistent with the input-specific formulation of MIDAS, where similar features can induce similar architecture parameters.
Although we do not explore this in our work, this approach could potentially be used to fine-tune architectures for improved performance on more challenging classes.

%% file: sec/5_conclusions.tex
\section{Conclusions}\label{section:conclusions}

We introduced MIDAS, a differentiable NAS method that replaces static architecture scalar parameters with input-specific parameters computed via patchwise self-attention, turning architecture selection into an adaptive, local decision. Coupled with a parameter-free, topology-aware search space, MIDAS achieves competitive results across several differentiable NAS benchmarks.

On NAS-Bench-201, MIDAS consistently discovers the optimal or near-optimal architecture. In the DARTS space, it reaches $97.42\%$ top-1 on CIFAR-10 and $83.38\%$ on CIFAR-100, and transfers to ImageNet with $75.4\%$ top-1.
On the RDARTS benchmarks (S1–S4), MIDAS remains competitive across all four spaces and achieves the best CIFAR-10 results among the compared methods on S2 and S4, with $2.49\%$ and $2.59\%$ error, respectively.

By modernizing DARTS with input-specific self-attention, MIDAS extends input-specific differentiable NAS with local architecture and topology-aware search. Future work will focus on refining decoding and extending MIDAS to new search spaces and tasks.

\section*{Acknowledgments}
We would like to thank Wojciech Jablonski for our prior joint work on differentiable neural architecture search and for providing the optimized DARTS reproduction that served as the starting point for our implementation. We also thank Piotr Bilinski for our prior joint work in this area.

%% file: sec/6_appendix.tex
\appendix
\onecolumn

\section{Use of Generative AI}

We used an LLM to assist in writing the manuscript and to generate plots. All ideas and experiments were designed and executed by the authors.

\section{Experiment Details}
\label{app:expdetails}

Table~\ref{table:expdetails} reports per-seed search validation and retraining results across DARTS, NAS-Bench-201, and RDARTS (S1–S4). For NAS-Bench-201, validation/test metrics are taken from the official tabular benchmark and we do not perform any additional retraining.

\begin{table}[t]
\caption{Detailed results of our experiments. \textbf{Search Val} is supernet validation accuracy at the end of search. \textbf{Retrain} reports params and validation accuracy after retraining the decoded architecture. NAS-Bench-201 validation/test values come from the official tabular benchmark.}
\label{table:expdetails}
\begin{center}
\begin{adjustbox}{width=\textwidth}
\begin{small}
\begin{sc}
\begin{tabular}{cccccccc}
\toprule
\multirow{2}{*}{\textbf{Search Space}} & \multirow{2}{*}{\textbf{Search Dataset}} & \multirow{2}{*}{\textbf{Seed}} & \multirow{2}{*}{\textbf{Search Val ($\%$)}} & \multicolumn{3}{c}{\textbf{Retrain}} \\
\cmidrule{5-7}
& & & & Params (M) & C10 Val ($\%$) & C100 Val ($\%$) \\
\midrule
\multirow{8}{*}{DARTS} & \multirow{4}{*}{CIFAR-10} & 1 & 89.51 & $-$ & $-$ & $-$ \\
& & 2 & 89.35 & $-$ & $-$ & $-$ \\
& & 3 & 89.59 & $-$ & $-$ & $-$ \\
& & 4 & 90.07 & 3.5 & 97.42 $\pm$ 0.06 & 83.38 $\pm$ 0.40 \\
\cmidrule{2-7}
 & \multirow{4}{*}{CIFAR-100} & 1 & 63.57 & $-$ & $-$ & $-$ \\
& & 2 & 64.80 & $-$ & $-$ & $-$ \\
& & 3 & 65.55 & 3.3 & 97.25 $\pm$ 0.09 & 82.67 $\pm$ 0.20 \\
& & 4 & 64.22 & $-$ & $-$ & $-$ \\

\midrule

\multirow{8}{*}{NAS-Bench-201} & \multirow{4}{*}{CIFAR-10} & 1 & 80.10 & $-$ & 91.55 & 73.49 \\
& & 2 & 79.24 & $-$ & 91.55 & 73.49 \\
& & 3 & 79.21 & $-$ & 90.20 & 70.71 \\
& & 4 & 80.76 & $-$ & 91.55 & 73.49 \\
\cmidrule{2-7}
 & \multirow{4}{*}{CIFAR-100} & 1 & 45.22 & $-$ & 91.21 & 71.60 \\
& & 2 & 45.67 & $-$ & 87.30 & 67.17 \\
& & 3 & 45.70 & $-$ & 91.08 & 71.00 \\
& & 4 & 44.90 & $-$ & 91.21 & 71.60 \\

\midrule

\multirow{4}{*}{S1} & \multirow{4}{*}{CIFAR-10} & 1 & 89.07 & 3.1 & 97.19 & 76.50 \\
&& 2 & 88.53 & $-$ & $-$ & $-$ \\
&& 3 & 88.91 & $-$ & $-$ & $-$ \\
&& 4 & 88.93 & $-$ & $-$ & $-$ \\

\midrule

\multirow{4}{*}{S2} & \multirow{4}{*}{CIFAR-10} & 1 & 89.81 & 3.9 & 97.51 & 77.81 \\
&& 2 & 89.30 & $-$ & $-$ & $-$ \\
&& 3 & 89.34 & $-$ & $-$ & $-$ \\
&& 4 & 89.51 & $-$ & $-$ & $-$ \\

\midrule

\multirow{4}{*}{S4} & \multirow{4}{*}{CIFAR-10} & 1 & 84.82 &  $-$ & $-$ & $-$ \\
&& 2 & 84.36 &  $-$ & $-$ & $-$ \\
&& 3 & 85.06 &  $-$ & $-$ & $-$ \\
&& 4 & 85.61 & 4.4 & 97.41 & 79.05 \\

\bottomrule
\end{tabular}
\end{sc}
\end{small}
\end{adjustbox}
\end{center}
\vskip -0.1in
\end{table}

\begin{figure}
  \centering
  \includegraphics[width=0.5\linewidth]{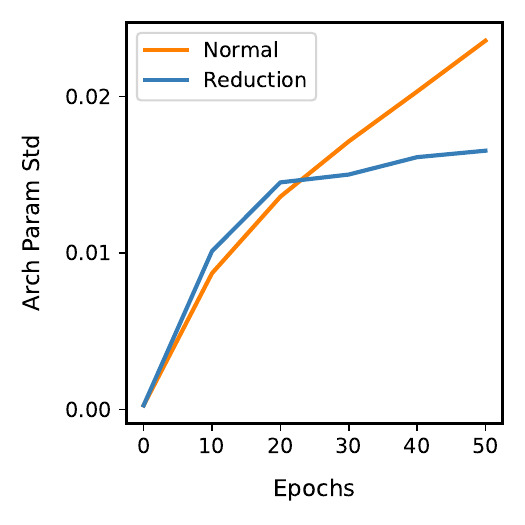}
  \caption{Standard deviation of input-specific architecture parameters $p^{(i, j)}$, averaged over candidate edges, nodes, and four seeds.}
  \label{fig:variance}
\end{figure}

\section{Searching and Retraining Protocols}\label{app:protocols}

We keep DARTS~\citep{darts} default hyperparameters except for three changes during search: (i) weight decay $3\times10^{-3}$ (vs.\ $3\times10^{-4}$), (ii) learning rate $10^{-4}$ for attention-based architecture parameters (vs.\ $10^{-3}$), (iii) cutout size $16$. All other settings remain consistent with DARTS and are summarized below. We control for these changes in \Cref{subsection:patch} by applying the same search hyperparameters to vanilla DARTS on NAS-Bench-201.

\subsection{Searching Protocol}

During the search phase, DARTS trains a supernet consisting of eight cells, each having four blocks (in the DARTS or RDARTS spaces) or three blocks (in the NAS-Bench-201 space). The second and fifth cells are reduction cells, which downsample the resolution of the activation map. The remaining cells are normal cells. Although standard DARTS employs a single shared normal cell and a single shared reduction cell architecture, our approach computes an input-specific architecture for each cell. The search space includes average pooling~(3$\times$3), max pooling~(3$\times$3), skip connection, depthwise separable convolution~(3$\times$3), depthwise separable convolution~(5$\times$5), dilated convolution~(3$\times$3), and dilated convolution~(5$\times$5). We omit the \textit{Zero} operation, as it is unnecessary in our topology-aware search space.

We search for 50 epochs with a batch size of 64 and $F=16$ channels, following the bilevel optimization scheme introduced by DARTS~\citep{darts}, which alternates gradient updates between convolution weights and architecture parameters. These updates are performed on disjoint halves of the training set to improve generalization. Convolution weights are trained using SGD with momentum~$0.9$, weight decay~$3\times10^{-3}$, and a learning rate following a cosine schedule from $0.025$ down to $10^{-3}$. Architecture parameters are trained with Adam (standard momentum), weight decay~$10^{-3}$, and a learning rate~$10^{-4}$. For data augmentation, we employ cutout of size 16. We partition the activation map into patches of size $8\times8$ (for DARTS and S1-S4) or $4\times4$ (for NAS-Bench-201).

\begin{figure}
    \centering
    \includegraphics[width=\linewidth]{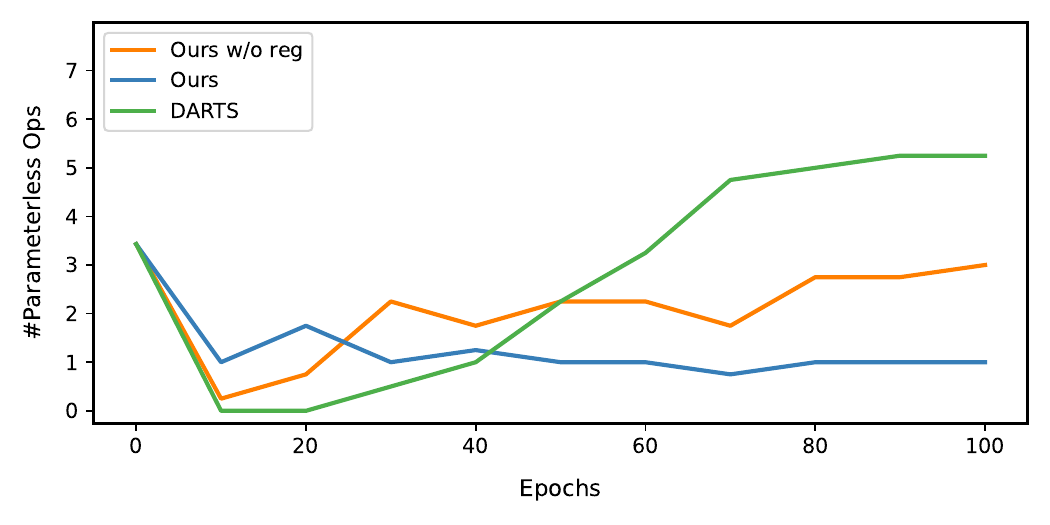}
    \caption{Number of non-learnable operations in the decoded normal cell during search (100 epochs).}
    \label{fig:skip}
\end{figure}

\subsection{Retraining Protocol}

After decoding the final architecture (detailed in Section~\ref{app:decoding}), we retrain decoded architectures with 20 cells, batch size of 96 and $F=36$ filters. The only exception is the $3\text{N}+2\text{R}$ variant in \Cref{table:cifar}, which - to ensure fair comparison and that the model has the same number of parameters as other methods - uses $F=44$.
Training runs for 600 epochs using SGD with momentum~$0.9$, weight decay~$3\times10^{-4}$, and a learning rate following a cosine schedule from $0.025$ down to $0$. For data augmentation, we employ cutout of size 16. In line with previous works, we apply drop path, increasing it linearly from $0$ to $0.2$, and auxiliary towers with weight~$0.4$. We report the mean of five independent runs per architecture and dataset. For ImageNet retraining, we rely on the SDARTS~\citep{sdarts} implementation and refer readers to their paper for additional details.

\section{Stability of Input-Specific Parameters}\label{app:robustness}

We next examine whether replacing scalar \(\alpha\) with input-specific architecture parameters affects the collapse behavior observed in DARTS. We track the number of non-learnable operations in the decoded normal cell over $100$ search epochs (to amplify the instability). Architecture parameters are averaged across normal cells to ensure fair comparison. As shown in~\Cref{fig:skip}, DARTS collapses, producing many skip connections, whereas MIDAS remains stable. Weaker regularization introduces more instability than in our default setting, but does not trigger an extreme collapse. Also, this mainly affects pooling which, unlike skip connections, can extract useful high-level features. Nonetheless, we apply stronger regularization to alleviate this behaviour.

\section{Choice of Patch Size}\label{app:ps}

We keep patch size (PS) fixed across spatial resolutions and restrict the practical candidates to \(PS\in\{4,8\}\). Both values divide all CIFAR feature-map sizes \((32,16,8)\). Smaller patches can downweight operations with larger receptive fields, while very large patches yield overly coarse features and weak discrimination among operations. We use \(PS=8\) for DARTS and RDARTS, and \(PS=4\) for NAS-Bench-201, as described in \Cref{subsection:expdetails}.

\section{Decoding Final Architecture}\label{app:decoding}

\subsection{Data Split Used for Decoding}

We compute input-specific parameters for each candidate edge pair and average over a randomly chosen $10\%$ subset of the training data. To test sensitivity to the subset used for decoding, we conduct an additional experiment with $45\%$, $45\%$, and $10\%$ training-set splits. The first two splits are used for weight and architecture training. The remaining $10\%$ is held out from both optimization procedures and used only for decoding. We then compare architectures decoded using:

\begin{itemize}
    \item The $45\%$ split used to optimize network weights
    \item The $45\%$ split used to optimize architecture parameters
    \item The $10\%$ held-out split
\end{itemize}

The three decoded genotypes are identical across all five cells except for a single operation in the final normal cell, where one edge in the third node alternates between a \(3\times3\) and \(5\times5\) dilated convolution. Thus, the decoded architecture is largely insensitive to the subset used to estimate the averaged architecture parameters. We therefore reuse DARTS’s $50\%/50\%$ split without a separate decoding split.

\subsection{Decoding Approaches}

We select the top-two candidate edges per node using marginalized input-specific parameters averaged over samples. We consider two approaches: (i) average parameters across all normal (respectively reduction) cells to obtain a single genotype for fair comparison, and (ii) AGNAS-style decoding - keep cells \#0, \#2, \#3, \#5, \#6 and stack as $(6\times c_{0}),\,c_{2},\,(6\times c_{3}),\,c_{5},\,(6\times c_{6})$. We intentionally avoid excessive tuning to reduce overfitting. In \Cref{fig:c10_cell,fig:c100_cell}, we visualize the decoded cells for the best CIFAR-10 and CIFAR-100 supernets, respectively.

\begin{figure}[h]
    \centering
    \begin{minipage}{0.32\textwidth}
        \centering
        \includegraphics[width=\linewidth]{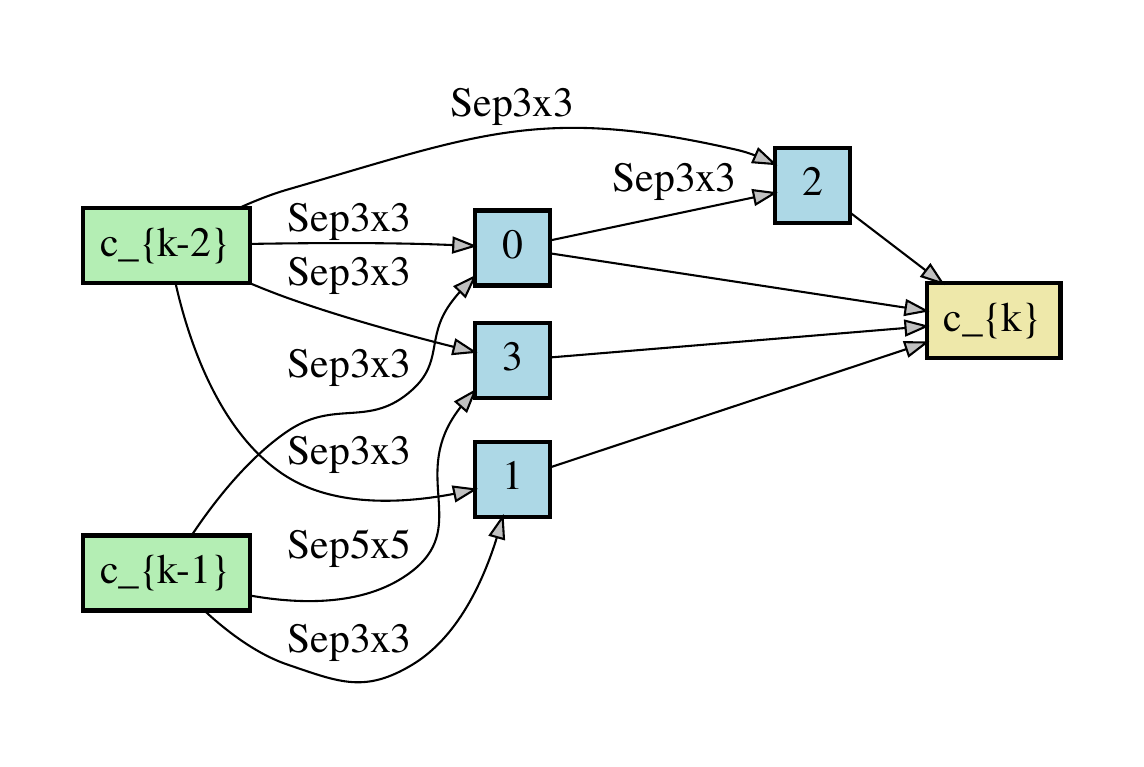}
        \captionsetup{labelformat=empty}
        \caption{(a) Cell 0 (normal)}
    \end{minipage}
    \hfill
    \begin{minipage}{0.32\textwidth}
        \centering
        \includegraphics[width=\linewidth]{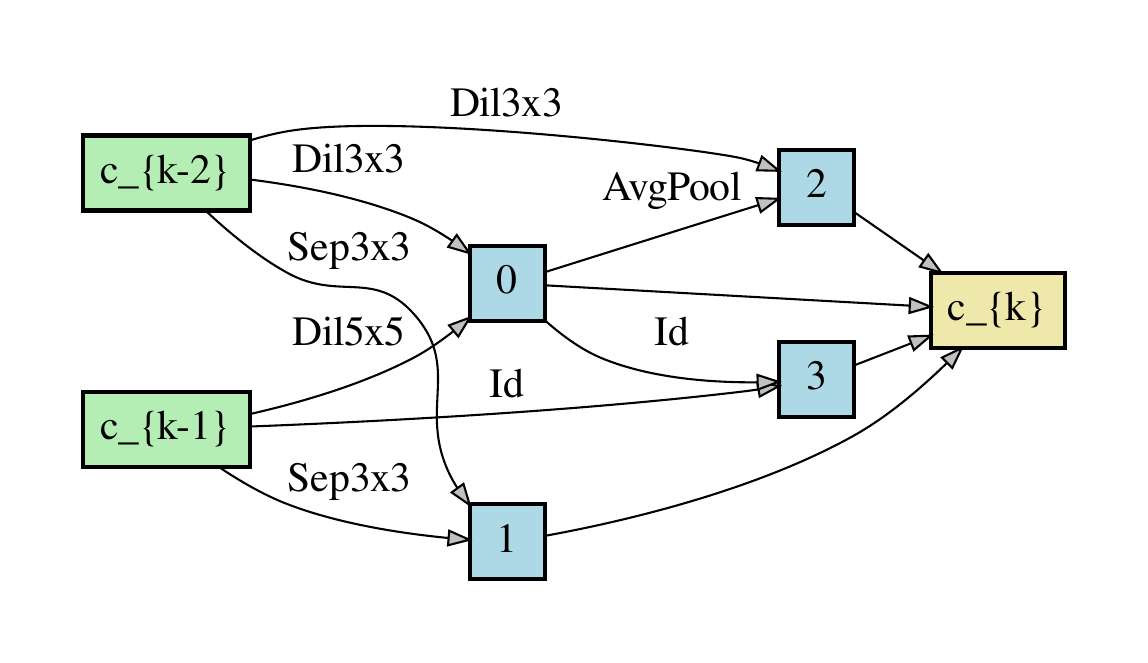}
        \captionsetup{labelformat=empty}
        \caption{(b) Cell 3 (normal)}
    \end{minipage}
    \hfill
    \begin{minipage}{0.32\textwidth}
        \centering
        \includegraphics[width=\linewidth]{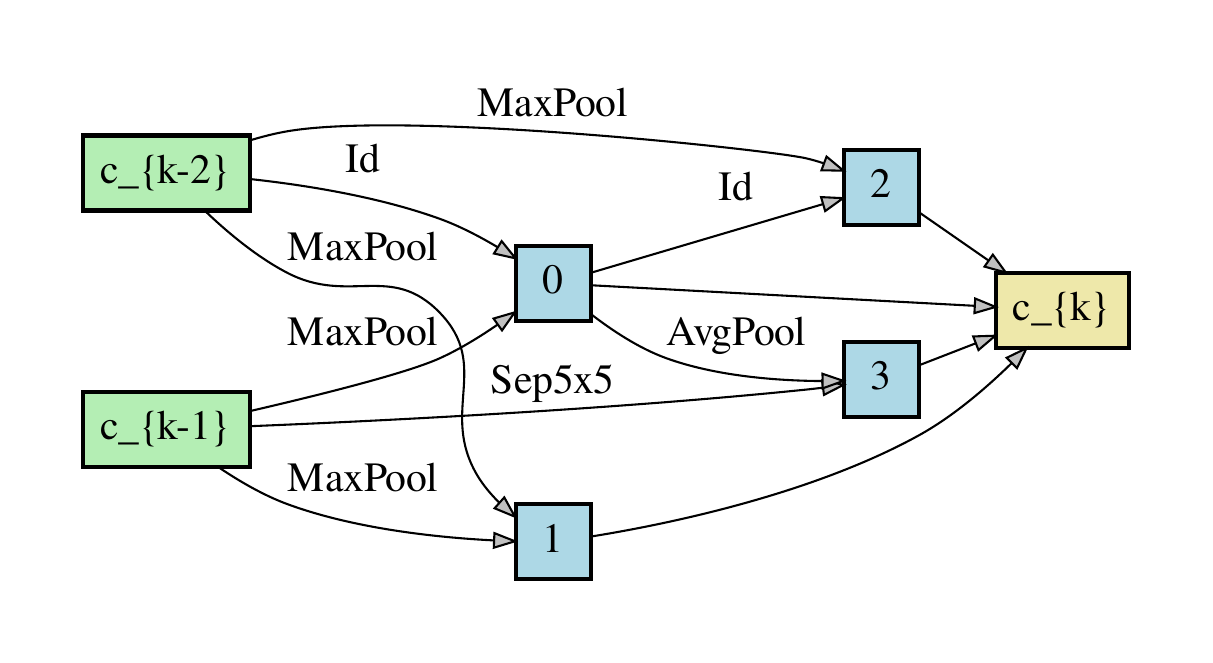}        \captionsetup{labelformat=empty}
        \caption{(c) Cell 6 (normal)}
    \end{minipage}

    \vspace{0.5cm}

    \begin{minipage}{0.32\textwidth}
        \centering
        \includegraphics[width=\linewidth]{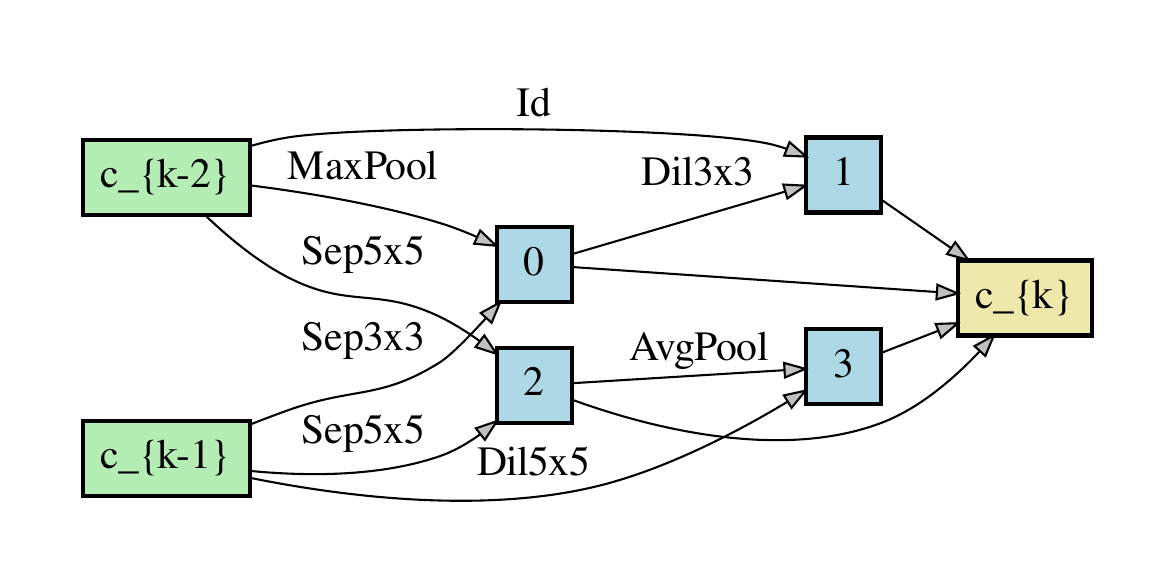}
        \captionsetup{labelformat=empty}
        \caption{(d) Cell 2 (reduction)}
    \end{minipage}
    \hspace{1cm}
    \begin{minipage}{0.32\textwidth}
        \centering
        \includegraphics[width=\linewidth]{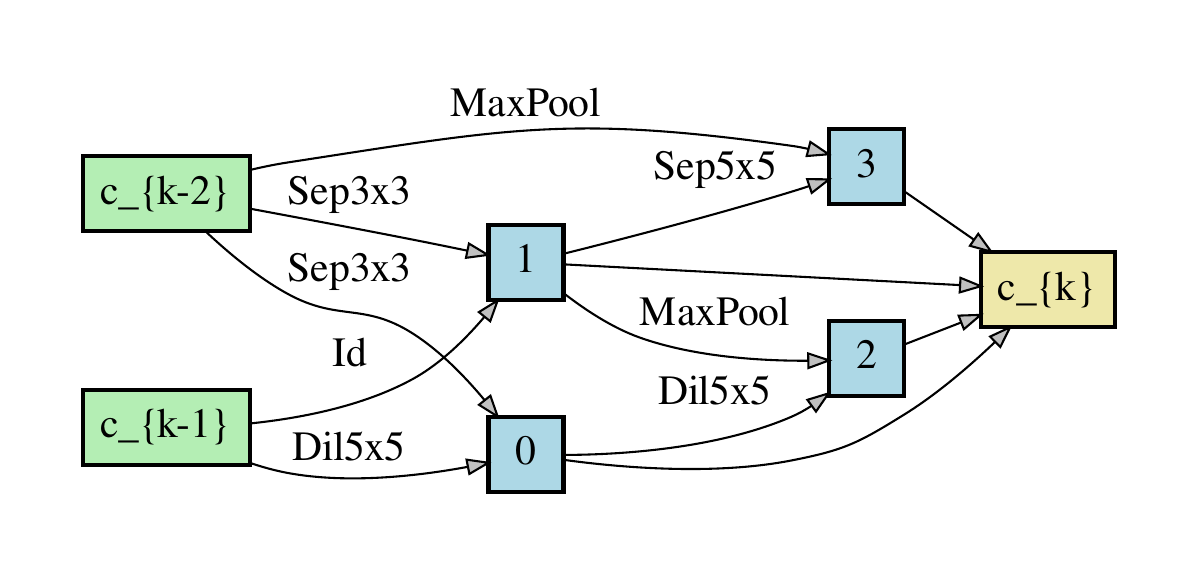}
        \captionsetup{labelformat=empty}
        \caption{(e) Cell 5 (reduction)}
    \end{minipage}

    \addtocounter{figure}{-1}\addtocounter{figure}{-1}\addtocounter{figure}{-1}\addtocounter{figure}{-1}\addtocounter{figure}{-1}

    \vskip 0.1in
    \caption{Architecture found by MIDAS on CIFAR-10.}
    \label{fig:c10_cell}
\end{figure}

\begin{figure}[h]
    \centering

    \begin{minipage}{0.32\textwidth}
        \centering
        \includegraphics[width=\linewidth]{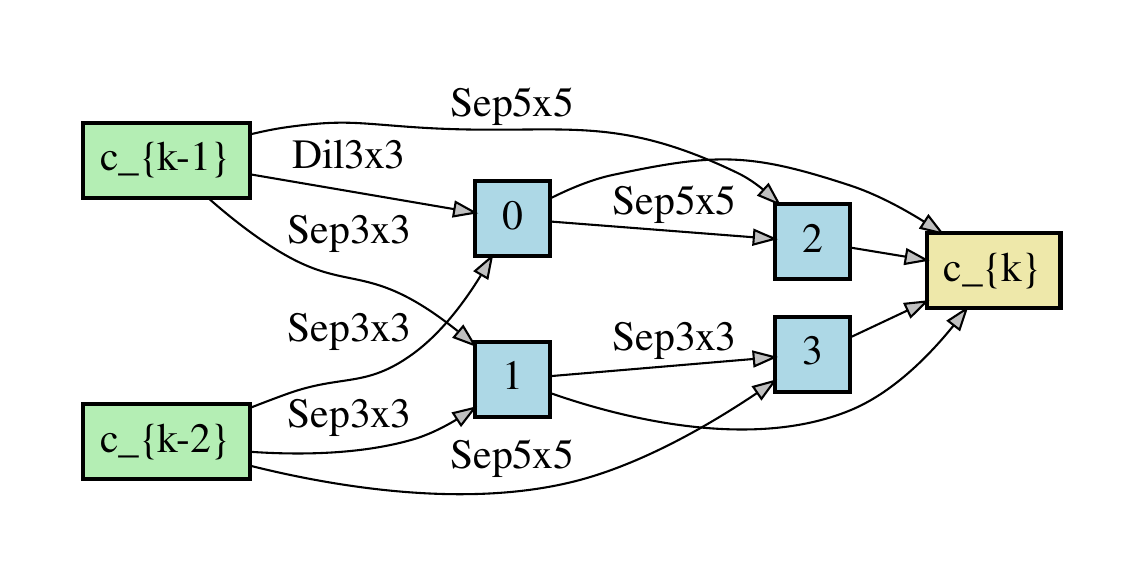}
        \captionsetup{labelformat=empty}
        \caption{(a) Cell 0 (normal)}
    \end{minipage}
    \hfill
    \begin{minipage}{0.32\textwidth}
        \centering
        \includegraphics[width=\linewidth]{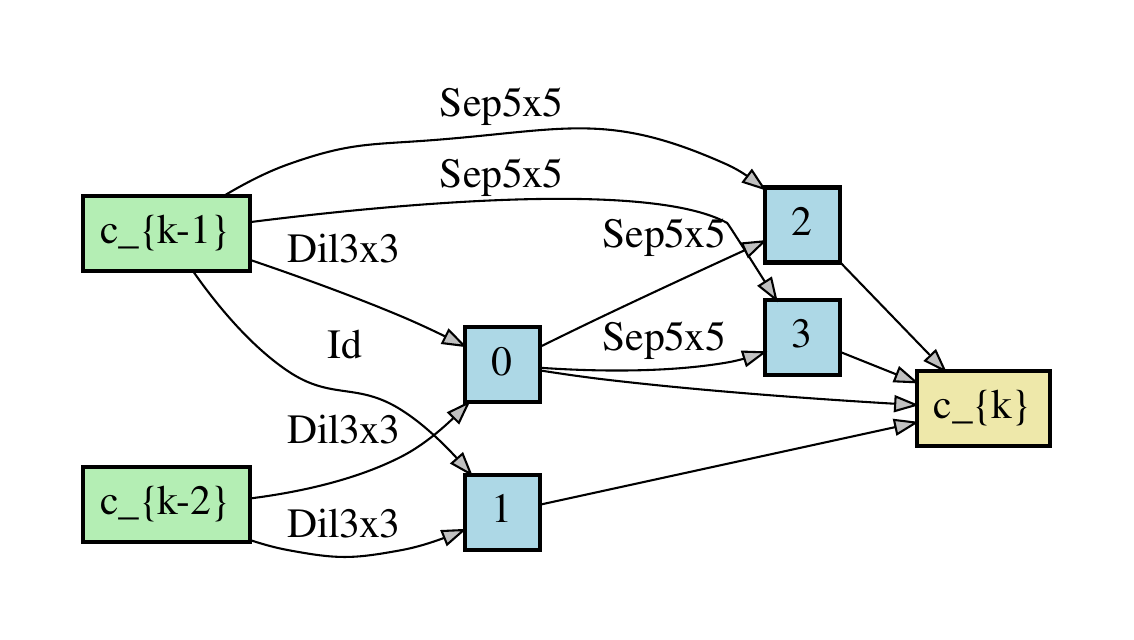}
        \captionsetup{labelformat=empty}
        \caption{(b) Cell 3 (normal)}
    \end{minipage}
    \hfill
    \begin{minipage}{0.32\textwidth}
        \centering
        \includegraphics[width=\linewidth]{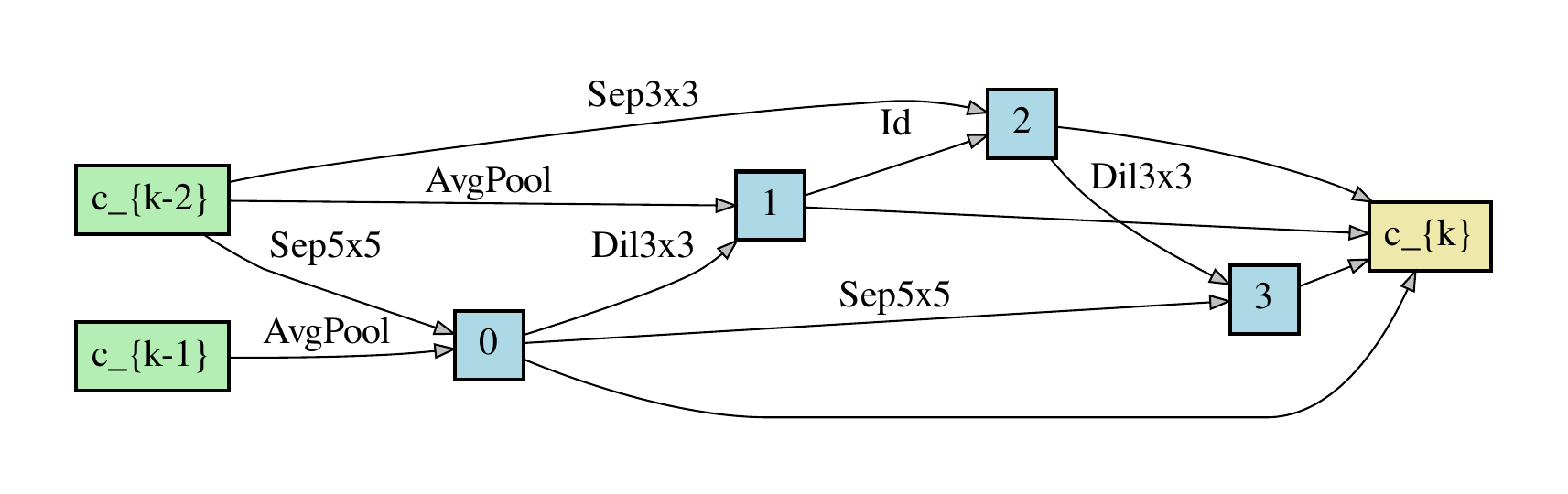}        \captionsetup{labelformat=empty}
        \caption{(c) Cell 6 (normal)}
    \end{minipage}

    \vspace{0.5cm}

    \begin{minipage}{0.32\textwidth}
        \centering
        \includegraphics[width=\linewidth]{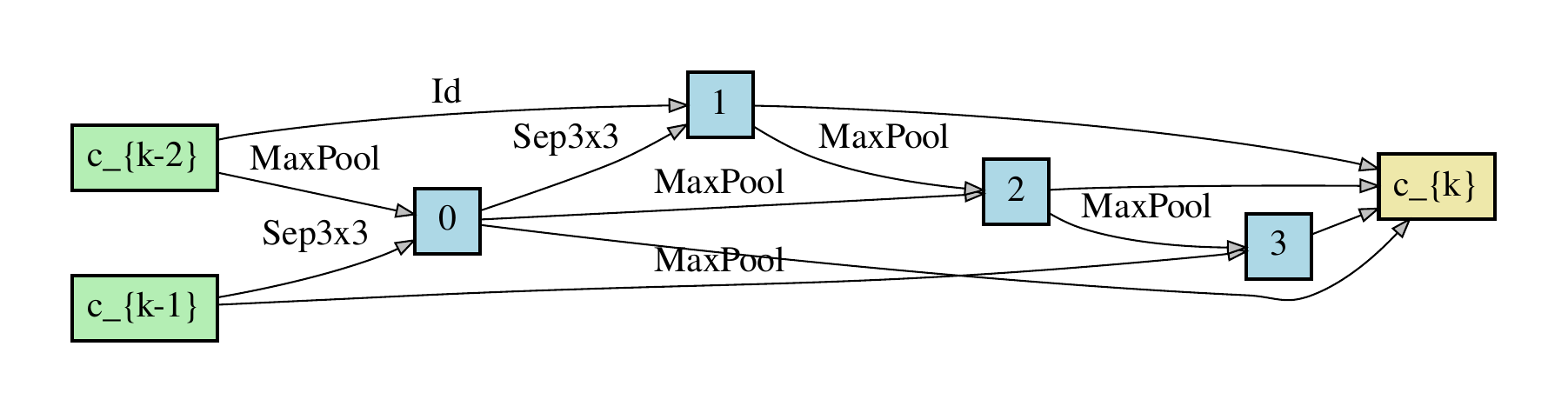}
        \captionsetup{labelformat=empty}
        \caption{(d) Cell 2 (reduction)}
    \end{minipage}
    \hspace{1cm}
    \begin{minipage}{0.32\textwidth}
        \centering
        \includegraphics[width=\linewidth]{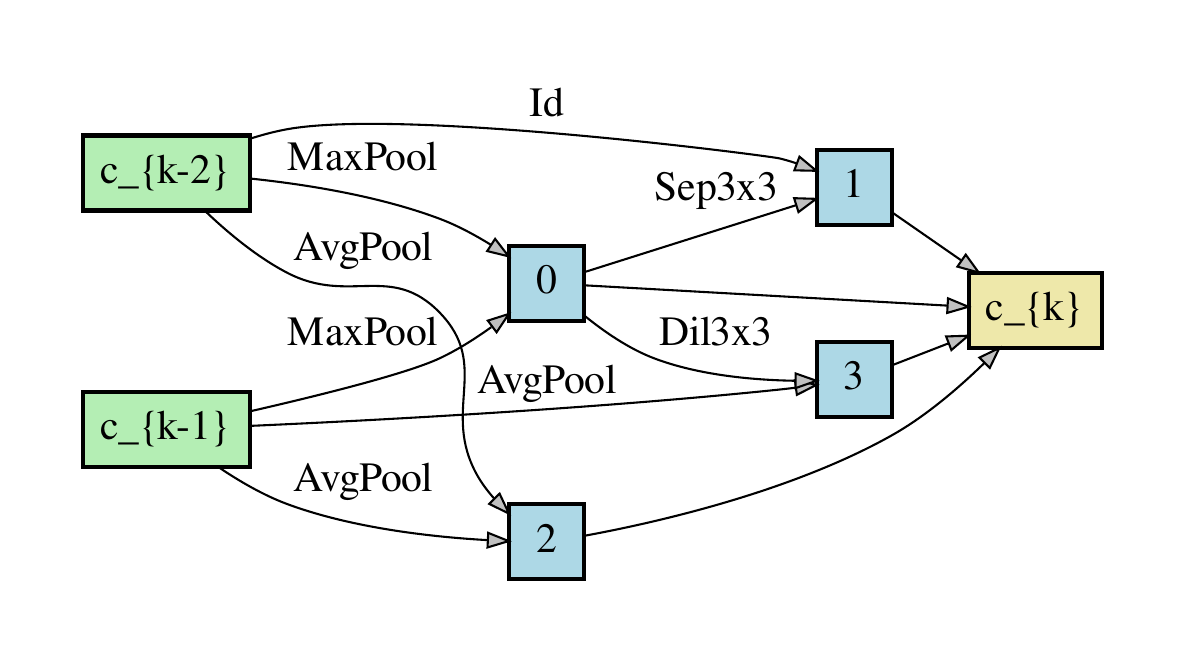}
        \captionsetup{labelformat=empty}
        \caption{(e) Cell 5 (reduction)}
    \end{minipage}

    \addtocounter{figure}{-1}\addtocounter{figure}{-1}\addtocounter{figure}{-1}\addtocounter{figure}{-1}\addtocounter{figure}{-1}

    \vskip 0.1in
    \caption{Architecture found by MIDAS on CIFAR-100.}
    \label{fig:c100_cell}
\end{figure}